\documentclass{article}
\usepackage{graphicx} 

\usepackage{microtype}
\usepackage{algorithm}
\usepackage{algorithmic}
\usepackage{subcaption}
\usepackage{booktabs} 
\usepackage[hidelinks]{hyperref}

\usepackage{amsmath, amsfonts, bm}
\usepackage{amssymb}
\usepackage{mathtools}
\usepackage{amsthm}
\usepackage{xcolor}
\usepackage[numbers, compress]{natbib}
\theoremstyle{plain}

\theoremstyle{definition}

\theoremstyle{remark}

\usepackage[nolist,nohyperlinks]{acronym}
\usepackage{multirow}

\newcommand{\R}{\mathbb{R}}
\newcommand{\N}{\mathbb{N}}
\newcommand{\C}{\mathbb{C}}

\newcommand{\seqlen}{L}
\newcommand{\modeldim}{D}
\newcommand{\headdim}{d_h}
\newcommand{\headkdim}{d_{\rm k}}
\newcommand{\headvdim}{d_{\rm v}}
\newcommand{\numheads}{H}
\newcommand{\nummodes}{M}
\newcommand{\numslots}{S}
\newcommand{\convkernel}{{\rm k}_{\rm conv}}

\newcommand{\key}{{\bm{k}}}
\newcommand{\query}{{\bm{q}}}

\newcommand{\Key}{{\bf{K}}}
\newcommand{\Query}{{\bf{Q}}}
\newcommand{\Val}{{\bf{V}}}
\newcommand{\Hid}{{\bf{X}}}

\newcommand{\Kcos}[1][]{{{\bf{K}}_{{#1}}^{\mathrm{cos}}}}
\newcommand{\Ksin}[1][]{{{\bf{K}}_{{#1}}^{\mathrm{sin}}}}
\newcommand{\Vcos}[1][]{{{\bf{V}}_{{#1}}^{\mathrm{cos}}}}
\newcommand{\Vsin}[1][]{{{\bf{V}}_{{#1}}^{\mathrm{sin}}}}

\newcommand{\cosmw}[1][]{\cos(m\omega{#1})}
\newcommand{\sinmw}[1][]{\sin(m\omega{#1})}
\newcommand{\eimw}[1][]{ e^{ \mathrm{i} m \omega {#1}} }

\title{Blurry Window Attention}
\author{Axel Laborieux$^1$ \and Christos Sourmpis$^1$ \and Juan Gabriel Kostelec$^1$ \and Qinghai Guo$^2$}
\date{\small $^1$Huawei, Zurich, Switzerland \\ 
      $^2$Huawei Advanced Computing and Storage Lab, Shenzhen, China}

\begin{document}

\begin{acronym}
  \acro{ssm}[SSM]{State-Space Model}
  \acro{la}[LA]{Linear Attention}
  \acro{bla}[BLA]{Blurry Window Attention}
  \acro{gla}[GLA]{Gated Linear Attention}
  \acro{gsa}[GSA]{Gated Slot Attention}
  \acro{abc}[ABC]{Attention with Bounded-memory Control}
  \acro{fla}[FLA]{Flash Linear Attention}
  \acro{gdn}[GDN]{Gated DeltaNet}
  \acro{swa}[SWA]{Sliding Window Attention}
  \acro{mqar}[MQAR]{Multi-Query Associate Recall}
  \acro{ai}[AI]{Artificial Intelligence}
  \acro{lm}[LM]{Language Model}
  \acro{rnn}[RNN]{Recurrent Neural Network}
\end{acronym}

\maketitle

\begin{abstract}
  The Softmax Attention operation in Transformer language models has a quadratic
  complexity in the sequence length and a growing state size in the form of KV cache,
  which becomes a bottleneck in long context scenarios.
  To overcome this limitation, alternative architectures with linear complexity 
  and finite state size have been introduced, such as \acp{ssm}, \ac{la}, and \ac{abc}.
  Though linear models achieve similar language perplexity as Transformers, they 
  are still behind in tasks which require retrieval or recall of specific information.
  In this work, we introduce \ac{bla} a novel \ac{abc} method inspired by \acp{ssm}.
  \ac{bla} stores a frequency window from which a blurry KV history is reconstructed via interpolation using Dirichlet kernels.
  \ac{bla} can be understood as a generalization of \ac{swa} depending on the Dirichlet kernels resolution or as a special case of the \ac{gsa}, where the decay factor is implemented with Dirichlet kernels.
  We describe in details the theory and efficient implementation of \ac{bla}. On the \ac{mqar} synthetic task, we show that the state efficiency of \ac{bla} is $8\times$ better than \ac{swa} and is competitive with popular linear attention models, and in the RegBench synthetic task, only \ac{bla} and \ac{swa} improve their performance as the state size grows among the linear models we tested.
\end{abstract}

\section{Introduction}

The Transformer architecture \citep{vaswani2017attention} and its attention mechanism
is one of the principal workhorses of large \acp{lm}.
The strength of attention comes from its ability to be parallelized over the sequence length 
and the all-to-all connection pathway between tokens, enabling direct interaction between distant time points.
However, the computing cost of this interaction is quadratic in the sequence length and becomes the main bottleneck when the context length
increases past the model dimension, which commonly occurs in scenarios such as agentic AI or long chain-of-thoughts.
In addition, Transformers require a growing KV cache during inference  
and each new token needs proportionally more compute.
\acf{swa} overcomes the quadratic complexity by truncating the KV history to a finite 
time window.
While stacking layers increases in principle the receptive field beyond the window size, the effect is not additive \cite{xiao2025sliding} and full attention layers are still required to maintain long range interaction.

To mitigate the quadratic complexity bottleneck of attention while still allowing for long
range performance, alternative architectures with linear sequence complexity have been designed.
The most prominent alternative architectures include \acfp{ssm} \citep{gu2020hippo, gu2021efficiently, gu2023mamba}, 
\acf{la} \citep{katharopoulos2020transformers, yang2023gated, yang2024gated}, and \acp{abc} \citep{peng2022abc, zhang2024gated}.
Like transformers, those neural networks are parallelizable over the sequence, but unlike transformers 
their linear sequence mixing operations use a finite state instead of a growing KV cache.
This gives linear \acp{lm} a computational advantage in long context scenario compared to transformers.
However, recent research points out that linear \acp{lm} fall short of attention variants in
specific tasks where long range information recall is needed \citep{bick2025understanding,von2025mesanet},
casting doubt on the long-term viability of purely linear \acp{lm} for text processing.

In this work, we present \acf{bla}, a novel linear attention architecture which is aimed at combining both the accurate retrieval of \ac{swa} and the long range dependencies of traditional \acp{ssm} and \ac{la} models.
While the state of linear attention stores key-value associations in an outer product format, \ac{bla} 
maintains separated key and value states, which makes \ac{bla} more similar to \ac{abc} and \ac{swa}.
However unlike \ac{abc} methods, \ac{bla} writing mechanism can be seen
as a generalization of \ac{swa}.
This is achieved by multiplying
and accumulating incoming keys and values independently across a finite set of Fourier modes similar to an \ac{ssm} like S4D \cite{gu2022parameterization}
Such a state space representation allows for
a lossy interpolation in the time domain up to a period using Dirichlet kernels. 
The current query is then used to compute softmax attention over the interpolated keys and values.
In the following, we first present the theory of \ac{bla} and then evaluate its performance on recall-intensive synthetic tasks.
We show that \ac{bla} has $8\times$ better state efficiency compared to \ac{swa} and comes close to popular linear models on the \ac{mqar} task.
In addition, \ac{bla} achieves similar performance to full attention on the RegBench task in contrast to \ac{gla} and \ac{gdn}, and is performing better than \ac{swa} for small state sizes. 

\section{Background}

We start by briefly recalling the operations of vanilla causal Softmax attention \citep{vaswani2017attention} 
and its linear variants, considering a single head and batch element for simplicity.

\subsection{Softmax Attention}
\label{sec:background_softmax_attn}

Given a sequence of $d$ dimensional vectors $\Hid \in \R^{\seqlen \times \modeldim}$ with 
sequence length $\seqlen$, Softmax attention projects the input to queries, keys and values sequences
 $\Query = \bm{W}_q \Hid, \Key = \bm{W}_k \Hid, \Val = \bm{W}_v \Hid \in \R^{\seqlen \times \modeldim}$ using projection matrices $\bm{W}_q, \bm{W}_k, \bm{W}_v \in \R^{\modeldim \times \modeldim}$.
The output is then given by the formula:
\begin{equation}
  {\bf{O}} = \mathrm{Softmax} \left( \frac{\Query \Key^\top}{\sqrt{\modeldim}} + {\bf{M}} \right)\Val \quad \in \R^{\seqlen \times \modeldim},
\end{equation}
where the softmax is applied row-wise. ${\bf{M}} \in \{-\infty, 0\}^{\seqlen \times \seqlen}$ is the causal mask
that prevents a query $\query_t$ from
querying future key vectors $\key_{t'>t}$.
The softmax term is a $\seqlen \times \seqlen$ matrix called the attention mask and is responsible for the $O(\seqlen^2 \modeldim)$ quadratic complexity in sequence length of vanilla Attention.
In the case of \acf{swa} with a window size $w$, the query $\query_t$ only attends to the keys of a sliding window $\key_{t'}$
where $t' \in [t-w, t]$, which brings the complexity to $O(\seqlen w \modeldim)$ at the cost of dropping long range interaction
between vectors.

\subsection{Attention with Bounded-memory Control}
\label{sec:background_abc}

\acf{abc} \cite{peng2022abc} introduces a cumulative softmax write gate \(\boldsymbol{\phi}_t\) that allows multiple tokens to be stored in a fixed‑size memory slot: 
\begin{equation}
    \widetilde{\mathbf{K}}_{t}=\widetilde{\mathbf{K}}_{t-1}+\boldsymbol{\phi}_{t}\otimes\mathbf{k}_{t},\quad \widetilde{\mathbf{V}}_{t}=\widetilde{\mathbf{V}}_{t-1}+\boldsymbol{\phi}_{t}\otimes\mathbf{v}_{t}. 
\end{equation}
\(\boldsymbol{\phi}_t\) is obtained via a normalized exponential of token features, giving a data dependent, FIFO‑like memory update while retaining the softmax attention over slots. This formulation can be expressed as a two‑pass linear attention, enabling hardware‑efficient chunkwise training with a small recurrent state.

\acf{gsa} \cite{zhang2024gated} builds on the ABC mechanism by adding a data‑dependent gating scalar \(\alpha_i \in [0,1]\) for each memory slot. At each step the key and value slots are updated with a gated recurrence
\begin{equation}
    \widetilde{\mathbf{K}}_t = \operatorname{Diag}(\boldsymbol{\alpha}_t)\widetilde{\mathbf{K}}_{t-1} + (1-\boldsymbol{\alpha}_t)\otimes \mathbf{k}_t
\end{equation}
(and analogously for \(\widetilde{\mathbf{V}}_t\)), which lets the model forget stale information and introduces a recency bias, addressing ABC’s inability to discard old tokens and its bias toward early tokens. This update can be written as a two‑pass Gated Linear Attention, enabling the same hardware‑efficient chunkwise training used for linear attention while providing a compact recurrent state and improved inference efficiency.

\subsection{State-Space Models}
\label{sec:background_ssm}

The \acf{ssm} literature can be traced back to the Legendre Memory Unit \cite{voelker2019legendre} and the Hippo theory \cite{gu2020hippo}.
The original question addressed by \acp{ssm} can be summarized as: given an incoming 1-D continuous signal $x(t)$ and a finite $N$ dimensional storage space, how to retain the most information about the signal? 
The \ac{ssm} theory shows that given some desired measure about the signal, we can project it on a basis to maintain a set of coordinates from which the signal can be approximated back.
Ignoring the step of discretization, the equations of a discrete \ac{ssm} are as follows:
\begin{equation}
  \label{eq:ssm}
  \begin{split}
    {\bm h}(t+1) & = A {\bm h}(t) + B x(t),  \\
    y(t) & = C {\bm h}(t) + D x(t).
  \end{split}
\end{equation}
Here ${\bm h}(t)$ is a $N$ dimensional state space representation of the signal $x(t)$.
The matrices $A, B, C, D$ are the parameters of the \ac{ssm}.
As we see from the equation, the state update is a linear recurrence, which allows for efficient parallelization over sequence length provided that the $A$ matrix is diagonalizable.
Early \acp{ssm} project the input signal on Legendre polynomials or Truncated Fourier modes \cite{voelker2019legendre, gu2020hippo}, which correspond to using specific parameter matrices.
Early \acp{ssm} and linear RNNs such as LRU or S4-FouT used to have $A$ matrices with complex eigenvalues \cite{orvieto2023resurrecting, gu2020hippo} for better expressivity since any real matrix $A$ is diagonalizable in $\C$ almost surely. 
This trend later changed for using diagonal real $A$ matrix learned from data, preceded by a short convolution to improve recall \cite{fu2022hungry, gu2022parameterization, gu2023mamba, de2024griffin}.
Interestingly, Mamba 3 comes back to using complex eigenvalues \cite{lahoti2026mamba}.

\section{Theory}

We now describe the theory of our novel \acf{bla} framework by first introducing it in a way that is similar to traditional \acp{ssm} and highlights the similarity of \ac{bla} with the \ac{ssm} literature.
We then show a more efficient implementation that does not require a convolution by exploiting the permutation invariance of softmax attention and resembles more \ac{abc}. 
Finally, we show how state decay similar to \ac{gsa} can be implemented in \ac{bla}, and make it look like a more general version of \ac{swa}.

\begin{figure*}[ht]
  \vskip 0.2in
  \begin{center}
    \centerline{\includegraphics[width=\textwidth]{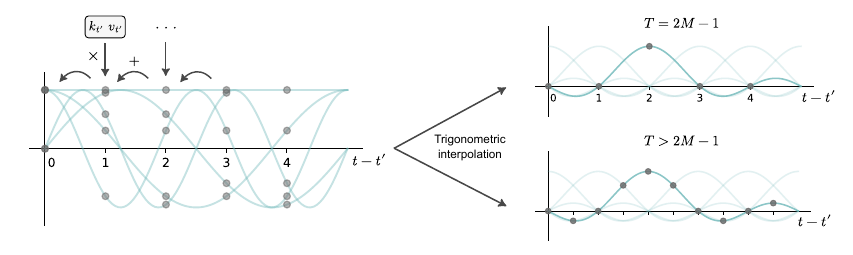}}
    \caption{
    Overview of the Blurry Window Attention mechanism.
    \textbf{Left:} The state of \ac{bla} is a convolution of the keys and values with the cosine
    and sine components of a set of $\nummodes$ Fourier modes parameterized by a period $T$.
    \textbf{Right:} In the specific case where $T=2M-1$, keys and values in the $(2M-1)$ time 
    window can be exactly recovered through trigonometric interpolation.
    When $T>2M-1$, the $2M-1$ keys and values are interpolated from the $T$ window.
    When the sequence exceeds $T$, the interpolated keys and values contain 
    anterior patterns due to periodicity.
    }
    \label{fig:cartoon}
  \end{center}
\end{figure*}

\subsection{Blurry Window Attention}
\label{sec:bla_theory_ssm}

Like other linear \acp{lm}, \acf{bla} maintains a finite state.
However, unlike \ac{la} variants which maintain an outer product state of keys and values,
\ac{bla} keeps separated key and value states 
$\Kcos, \Vcos \in \R^{\modeldim \times \nummodes}$ and 
$\Ksin, \Vsin \in \R^{\modeldim \times (\nummodes-1)}$ where $M$ 
encodes a number of Fourier modes and is the main parameter of \ac{bla}.
These states are initialized to zero.
Defining $\omega = 2\pi/(2M-1)$, the recurrent update of the key state is written as :
\begin{equation}
  \label{eq:recurrence}
  \begin{split}
  \Kcos[m,t+1] & = \cosmw \Kcos[m,t] - \sinmw \Ksin[m,t] + \key_t,  \\
  \Ksin[m,t+1] & = \sinmw \Kcos[m,t] + \cosmw \Ksin[m,t]
  \end{split}
\end{equation}
for $m \in [0,...,\nummodes-1]$. The value state is updated similarly.
These equations can be written more compactly in complex notation, and consists 
in linear recurrences with coefficients $\eimw$, which is similar to a diagonal \ac{ssm} with complex diagonal $A$ matrix, like S4D \cite{gu2020hippo, gu2022parameterization}, and a $B$ matrix full of ones (Eq.~\eqref{eq:ssm}).
We adopt real notation throughout
to closely follow how the algorithm is implemented in practice with real data types.
If we solve the recurrence in Eq.~\eqref{eq:recurrence}, we obtain the closed form formula for the states:
\begin{equation}
  \label{eq:states}
  \begin{split}
  \Kcos[m,t] & = \sum_{t'=1}^t \cosmw[(t-t')]~ \key_{t'},  \\
  \Ksin[m,t] & = \sum_{t'=1}^t \sinmw[(t-t')]~ \key_{t'}.
  \end{split}
\end{equation}
which are the keys convolved with the cosine and sine functions.
We defer the derivation to Appendix~\ref{sec:app_proofs}.
The left plot in Figure~\ref{fig:cartoon} illustrates the convolution of the keys and values with the trigonometric functions. 
By combining the two states with the appropriate coefficients, we can propagate the memory of the keys and values with an arbitrary function into the future. 
In particular, we consider a $(2\nummodes-1)$-periodic continuous function $f$ with highest mode $(M-1)\omega$. 
We can write $f$ in terms of its coordinates on the Fourier basis functions
$(t \mapsto 1, t \mapsto \cosmw[t], t \mapsto \sinmw[t], ~ m<M)$:
\begin{equation}
  \label{eq:f}
  f(t) = \sum_{m=0}^{\nummodes-1} a_m \cos(m \omega t) + b_m \sin (m \omega t),
\end{equation}
setting $b_0=0$ by convention.
At a time $t$, even if we do not have access to past keys and values for $t'<t$,
 we can write a convolution of past keys and values with any such $f$ function:
\begin{equation}
  \label{eq:conv}
  \begin{split}
  (f \ast \key)(t) &= \sum_{t'=1}^t f(t-t')~ \key_{t'} \\ 
  &= \sum_{t'=1}^t \sum_{m=0}^{\nummodes-1} \left( a_m \cos(m \omega (t-t')) + b_m \sin (m \omega (t-t')) \right)~ \key_{t'} \\ 
  &=  \sum_{m=0}^{\nummodes-1} a_m \sum_{t'=1}^t\cos(m \omega (t-t'))\key_{t'} + b_m  \sum_{t'=1}^t\sin (m \omega (t-t'))\key_{t'} \\ 
  &= \sum_{m=0}^{\nummodes-1}a_m \Kcos[m,t] + b_m \Ksin[m,t] \\
  &= \Kcos[t]\cdot {\bf{a}} + \Ksin[t] \cdot {\bf{b}},
  \end{split}
\end{equation}
where ${\bf{a}} = (a_0, ..., a_{M-1})^\top$ and ${\bf{b}} = (b_0, ..., b_{M-1})^\top$.
We used the state formula of Eq.~\eqref{eq:states} and exchanged the sums over time and over modes.
In the \ac{ssm} formalism, this step corresponds to the state readout using the $C$ matrix (Eq.~\eqref{eq:ssm}).
Looking back at Eq.~\eqref{eq:conv}, one interesting $f$ function to consider is
 the so-called Dirichlet kernel \cite{edwards1979dirichlet}:
\begin{equation}
  \label{eq:dirichlet}
   D_{M}(t) = \frac{1}{2M-1} + \frac{2}{2M-1}\sum_{m=1}^{M-1} \cosmw[t].
 \end{equation}
 This corresponds to setting $a_0 = 1/(2M-1)$, $a_m = 2/(2M-1) ~m>1$ and $b_m =0 ~ \forall m$.
 $D_M$ is such that $D_M(0)=1$ , $D_M (t) = 0$ for integers $t \in [1, ... 2M-2]$
  (Fig.~\ref{fig:cartoon} top-right).
Applying the result of Eq.~\eqref{eq:conv} with those coefficients gives:
\begin{equation}
    \begin{split}
        (D_M \ast \key)(t) &= \sum_{t'=1}^t D_M(t-t') ~ \key_{t'} \\ 
        &= \sum_{t'=1}^t \delta(t \equiv t' ~[2M-1]) ~ \key_{t'} \\ 
        &= \sum_{t' \equiv t ~[2M-1]}  \key_{t'}.  
    \end{split}
  \label{eq:conv_key}
\end{equation}
In particular if the sequence is less than $2M-1$ in length it gives the latest key.
We can also consider the translated Dirichlet kernels to any $\Delta t$ integers in $[0, ..., 2M-2]$:
\begin{equation}
    \label{eq:translated_dirichelt}
    D_{M, \Delta t}(t) = D_M(t-\Delta t)
\end{equation}
We show in Appendix \ref{sec:app_proofs} that the Fourier coefficients of the $D_{M, \Delta t}$ are:
\begin{align}
\label{eq:translated}
&{\bf{A}}_{\Delta t,0} = \frac{1}{2M-1}, &&{\bf{B}}_{\Delta t,0} = 0, \nonumber \\
 &{\bf{A}}_{\Delta t,m>0} = \frac{2\cosmw[\Delta t]}{2M-1}, &&{\bf{B}}_{\Delta t,m>0} = \frac{2\sinmw[\Delta t]}{2M-1}.
\end{align}
We gather those coefficients in two matrices ${\bf{A}}, {\bf{B}} \in \R^{(2M-1)\times M}$.
We can then define $\tilde{\bf{K}}_t \in \R^{\modeldim \times (2\nummodes-1)}$ (respectively $\tilde{\bf{V}}_t$) as:
\begin{equation}
  \label{eq:compressed}
  \tilde{{\bf{K}}}_t =  \Kcos[t]{\bf{A}}^{\top} +  \Ksin[t] {\bf{B}}^{\top}.
\end{equation}
When the sequence length is less than $2M-1$, $\tilde{\bf{K}}_t$ contains the keys of the previous $2M-1$ time window.
When the sequence length exceeds $2M-1$, the keys are added modulo $2M-1$ following Eq.~\eqref{eq:conv_key}.
More generally, we can set $\omega = 2\pi/T$, where $T$ ($T \geq 2\nummodes-1$) is the period of the fundamental frequency and is the second most important parameter of our model.
This interpolates keys and values instead of indexing them, which effectively blurs the sequence (Fig.~\ref{fig:cartoon} bottom right).
The output of \acf{bla} is computed as:
\begin{equation}
  \label{eq:output}
  {\bf{O}} = \mathrm{Softmax}\left( \frac{\Query \tilde{\bf{K}}^\top}{\sqrt{\modeldim}} + {\bf{M}_{BLA}} \right)\tilde{\Val}.
\end{equation}
The mask ${\bf{M}_{BLA}}$ is not added for enforcing causality, but to prevent the model from attending to the zero-valued keys and values before the beginning of the sequence.
The reason is that what the model sees is the sequence of 
keys and values from the present time to the past, so  
keys and values before the beginning of the sequence should not be attended to.

\begin{figure}[ht]
  \vskip 0.2in
  \begin{center}
    \centerline{\includegraphics[width=\columnwidth]{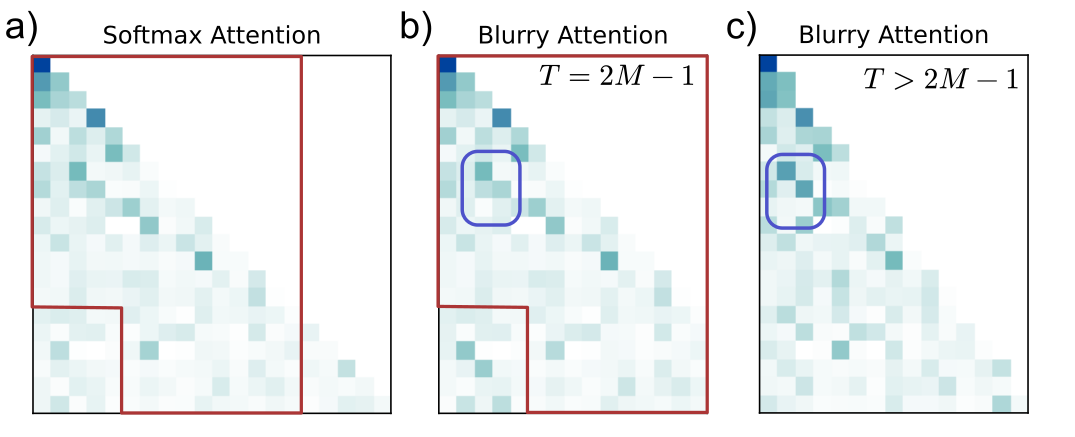}}
    \caption{
    Comparison of Vanilla Attention and Blurry Attention.
    \textbf{a)} Vanilla Attention mask with causal masking.
    \textbf{b)} Blurry attention mask when $T=L=2m-1$ is identical to vanilla attention due to exact interpolation (compare red boxes).
    When $L>T=2m-1$, the \ac{bla} mask becomes oblong and contains superposition of KV modulo $T$ for $t \in [T,L]$.
    \textbf{c)} When $L=T>2m-1$, the mask has lower token resolution (compare blue boxes).
    }
    \label{fig:attn_mask}
  \end{center}
\end{figure}

In Fig.~\ref{fig:attn_mask}, we compare the attention masks of full attention and \ac{bla} obtained with Eq.~\eqref{eq:output} for $M=8$, in the cases of choosing $T=2M-1$ and $T>2M-1$.
We see that \ac{bla}, in contrast to many other \acp{ssm} and similar to \ac{abc} methods, can exhibit ``sharp'' attention matrices, which is important for retrieval \cite{zhang2024hedgehog}.
When $T=2M-1$ and the sequence length is smaller than $2M-1$ and $T=2M-1$, \ac{bla} has the same attention mask as vanilla attention (red boxes in Fig.~\ref{fig:attn_mask}a,b).
However when the sequence length exceeds $2M-1$, the tokens from previous windows are summed modulo $2\nummodes-1$ instead of dropped like in \ac{swa} (bottom left corner in Fig.~\ref{fig:attn_mask}a,b).
This in theory allows \ac{bla} to capture longer range dependencies compared to \ac{swa}, but comes at the risk of having the state diverge, but we show later how a decay mechanism can be added to \ac{bla}.
Finally, when $T>2M-1$, the attention mask has different query and key time scales, leading to a blurry attention mask. 
\ac{bla} is also similar to the \ac{abc} model with $S=2M-1$ slots. However in contrast to \ac{abc} and due to its underpinning to Fourier theory the different modes/slots can be combined to generate an arbitrary function $f$ Eq.~\eqref{eq:f}.

\subsection{A More Efficient Implementation}
\label{sec:theory_bla_efficient}

In section \ref{sec:bla_theory_ssm}, we introduced \ac{bla} from the standpoint of the \ac{ssm} theory.
As a result, the state formula Eq.~\eqref{eq:states} has the form of a convolution.
We show here that owing to the permutation invariance of softmax attention we can instead write the state of \ac{bla} as a cumulative sum.
We begin by rewriting the blurred key state for a specific slot $\Delta t \in [0, ... 2M-2]$ at a time step $t$.
We have using Eqs.~\eqref{eq:conv}, \eqref{eq:conv_key}, and \eqref{eq:compressed}:
\begin{equation}
  \begin{split}
  \tilde{{\bf{K}}}_{t}[\Delta t] &=  \sum_{t'=1}^{t}  D_{M, \Delta t}(t-t') ~\key_{t'} = \sum_{t'=1}^{t}  D_{M, 0}(t-t'-\Delta t) ~\key_{t'}   \\
   &= \sum_{t'=1}^{t}  D_{M, 0}(t'-(t-\Delta t)) ~\key_{t'} = \sum_{t'=1}^{t}  D_{M, t-\Delta t}(t') ~\key_{t'} \\
   &= \sum_{t'=1}^{t}  D_{M, \Delta\tau}(t') ~\key_{t'},
  \end{split}
\end{equation}
where we use the parity of $D_{M,0}$ (Eq.~\eqref{eq:dirichlet}), and define $\Delta \tau = t-\Delta t~[2M-1]$.
Therefore we can use the opposite manipulation and find that:
\begin{equation}
  \begin{split}
  \tilde{{\bf{K}}}_{t}[\Delta \tau] &= \sum_{t'=1}^{t}  D_{M, \Delta t}(t') ~\key_{t'}.
  \label{eq:permuted}
  \end{split}
\end{equation}
As a result, we can simply compute first the quantity:
\begin{equation}
  \label{eq:compressed_simplified}
  \begin{split}
   D_{M, \Delta t}(t') = \cos(\omega t'{\bf{m}}){\bf{A}}^{\top} +  \sin(\omega t'{\bf{m}}) {\bf{B}}^{\top},
  \end{split}
\end{equation}
where ${\bf{m}} = (0, ..., M-1)$ and multiply the current key with it.
Accumulating this quantity over time is computing $\tilde{{\bf{K}}}_{t}[\Delta \tau]$ (Eq.~\eqref{eq:permuted}), which is a time-rolling column of $\tilde{{\bf{K}}}_{t}\textbf{}$.
Since the softmax attention is permutation invariant, the output is not changed and we can use $\tilde{{\bf{K}}}_{t}[\Delta \tau]$ instead of $\tilde{{\bf{K}}}_{t}[\Delta t]$.
With this manipulation, we see that \ac{bla} can be computed with a cumulative sum operation similar to linear attention.
We provide the efficient algorithm for the recurrent mode in Alg.~\ref{alg:recurrent_bla_swa} and the chunk mode in Alg.~\ref{alg:chunk_bla_swa} of Appendix \ref{sec:app_pseudo_code}.

\begin{algorithm}[tb]
\caption{Efficient recurrent \ac{bla}}
\label{alg:recurrent_bla_swa}
\begin{algorithmic}
\STATE {\bfseries Input:} $\Query, \Key, \Val \in \mathbb{R}^{\seqlen \times \numheads \times \modeldim}$, period $T \in \mathbb{R}^{\numheads}$,
interpolation matrices ${\bf{A}} \in \mathbb{R}^{2\nummodes-1 \times \nummodes}, {\bf{B}} \in \mathbb{R}^{2\nummodes-1 \times \nummodes}$
\STATE Clamp period: $T \gets \max(T, 2 \cdot \nummodes - 1)$
\STATE Compute dilated time grid:
\STATE $\mathrm{dilated\_time} \gets \mathrm{round}\left(\frac{[0,\dots,2\nummodes-1-1]}{2\nummodes-1} \otimes T\right)$
\STATE $\omega \gets 2\pi/T$
\STATE Compute modes: 
\STATE $\mathrm{modes} \gets \omega \otimes [0, 1, \dots, \nummodes-1]$
\STATE Initialize compressed KV states:
\STATE ${\bf{K}}_{\mathrm{prev}}, {\bf{V}}_{\mathrm{prev}} \gets {\bf{0}}^{\numheads \times \modeldim \times 2\nummodes-1}$ 
\STATE Initialize output: ${\bf{O}} \gets {\bf{0}}^{\seqlen \times \numheads \times \modeldim}$
\FOR{$t = 0$ to $\seqlen-1$}
\STATE Compute interpolation coefficients:
\STATE \quad $\mathrm{interpolate} \gets $
\STATE \qquad $\cos(t \cdot \mathrm{modes}) \cdot {\bf{A}}^\top + \sin(t \cdot \mathrm{modes}) \cdot {\bf{B}}^\top$
\STATE Update compressed KV states:
\STATE \quad ${\tilde{\bf{K}}} \gets {\bf{K}}_{\mathrm{prev}} + \Key[t] \otimes \mathrm{interpolate}$
\STATE \quad ${\tilde{\bf{V}}} \gets {\bf{V}}_{\mathrm{prev}} + \Val[t] \otimes \mathrm{interpolate}$
\STATE \quad ${\bf{K}}_{\mathrm{prev}} \gets {\tilde{\bf{K}}}$ 
\STATE \quad ${\bf{V}}_{\mathrm{prev}} \gets {\tilde{\bf{V}}}$
\STATE Compute attention weights:
\STATE \quad ${\bf{M}}_{\mathrm{BLA}} \gets \mathrm{where}(t  \geq \mathrm{dilated\_time}, 0.0, -\infty)$
\STATE \quad ${\bf{O}}[t] \gets  \mathrm{Softmax}\left(\frac{\Query[t] \cdot {\tilde{\bf{K}}}}{\sqrt{\modeldim}} + {\bf{M}}_{\mathrm{BLA}}\right) \cdot {\tilde{\bf{V}}}$
\ENDFOR
\STATE {\bfseries Return:} $\bf{O}$
\end{algorithmic}
\end{algorithm}

\subsection{Adding State Decay to \ac{bla}}
\label{sec:bla_decay}
One potential issue of \ac{bla} is that the states $\bf{\tilde\Key}$ and $\bf{\tilde\Val}$ continuously accumulate keys and values over time, as shown in Eqs.~\eqref{eq:recurrence} and \eqref{eq:permuted}.
For long sequences, this unbounded growth can lead to numerical instabilities.
A natural solution to mitigate this issue is to introduce a decay mechanism into the state update rule similarly to other \acp{ssm} \citep{gu2023mamba,yang2023gated}. A principled choice for this decay is to multiply the previous state by ($1 - D_{M, \Delta t}(t')$), as introduced in Eq.~\eqref{eq:compressed_simplified}. 
This formulation is particularly advantageous because it allows us to generalize \ac{swa}, and retrieve the classic \ac{swa} when the period $T$ is set to $2 \nummodes - 1$.
As established in Eq.~\eqref{eq:conv_key}, when the sequence length is less than or equal to $2 \nummodes - 1$, the matrices $\bf{\tilde\Key}$ and $\bf{\tilde\Val}$ contain the keys and values from the preceding $2 \nummodes - 1$ time steps.
By incorporating this decay, we can implement a controlled forgetting mechanism: it ``flushes'' the previous exact value when $T = 2M - 1$, similar to \ac{swa}, or maintains a decaying history of previous values when $T > 2M - 1$. 
By setting the decay term as $1 - D_{M, \Delta t}(t')$, we get almost the same formulation as \ac{gsa}, which allows us to re-use the efficient implementations of the FLA repository\footnote{Double GLA pass, and implemented in \hyperlink{https://github.com/fla-org/flash-linear-attention/blob/main/fla/ops/gsa/chunk.py}{https://github.com/fla-org/flash-linear-attention/blob/main/fla/ops/gsa/chunk.py}}.

\section{Experiments}

We perform experiments to evaluate how \ac{bla} performs in function of its state size.
We mainly compare \ac{bla} with popular linear models from the \ac{fla} repository \cite{yang2024fla}, namely \ac{gla} \cite{yang2023gated}, \ac{gdn} \cite{yang2024parallelizing}, \ac{gsa} \cite{zhang2024gated} and \ac{swa}.
For fair comparison, we match the state size of different models according to the state sizes in Table \ref{tab:statesize}.
We choose the \acf{mqar} \cite{arora2023zoology} and RegBench \cite{akyurek2024context} synthetic benchmarks to evaluate \ac{bla} because they require retrieval abilities.
The hyperparameters are given in Appendix \ref{sec:app_hparams}.

\begin{table}[t]
  \caption{The state sizes of different linear \acp{lm} in function of model parameters.}
  \label{tab:statesize}
  \begin{center}
    \begin{small}
      \begin{tabular}{lcccr}
        \toprule
        Model              & Parameter & State size (elements)     \\ 
        \midrule
        Short conv    & Kernel size $\convkernel$         & $\modeldim(\convkernel-1)$  \\ 
        \ac{swa}      & Window size $w$      & $\numheads (\headkdim+\headvdim) \min (w, \seqlen)$  \\
        \ac{gla}      & Head dim. $\headdim$ & $\numheads \headkdim \headvdim$  \\
        \ac{gdn}      & Head dim. $\headdim$ & $\numheads \headkdim \headvdim$  \\
        \ac{gsa}      & \# of slots $\numslots$ & $\numheads (\headkdim+\headvdim) \numslots$  \\
        \ac{bla}      & \# of modes $\nummodes$ & $\numheads (\headkdim+\headvdim) (2\nummodes -1)$  \\
        \bottomrule
      \end{tabular}
    \end{small}
  \end{center}
  \vskip -0.1in
\end{table}

\subsection{Multi Query Associative recall}

\begin{figure*}[ht]
  \vskip 0.2in
  \begin{center}
    \centerline{\includegraphics[width=\textwidth]{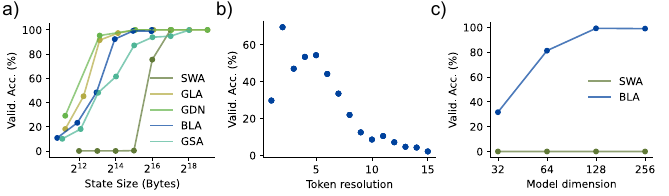}}
    \caption{
    Results on Multi-Query Associative recall \citep{arora2023zoology}.
    \textbf{a)} Pareto Frontier of \ac{bla} compared to other linear models. \ac{bla} improves the
    pareto frontier of \ac{swa} by $8\times$. 
    \textbf{b)} The period parameter controlling the token resolution has a high impact on performance, with
    a clear optimum at a resolution of 2 tokens.
    \textbf{c)} \ac{bla} can leverage bigger model dimensions to store more information and improve
    performance, while \ac{swa} cannot.
    }
    \label{fig:mqar}
  \end{center}
\end{figure*}

We first evaluate \ac{bla} without the decay mechanism on the \ac{mqar} task.
In this task, the model is presented with a sequence of key-value pairs and is trained to output the correct values of multiple keys.
We use a challenging setting of sequence length of 512 and 64 key value pairs.
We produce a Pareto frontier shown in Fig.~\ref{fig:mqar}a by measuring the maximum validation accuracy over a sweep of learning rates, seeds, and parameters controlling the state sizes (Table \ref{tab:statesize}).
We found that \ac{bla} uses its state size more efficiently than \ac{swa} (by $8 \times$) and \ac{gsa}, and comes close to \ac{gla} and \ac{gdn}.
We found that a short convolution was required for \ac{gla}, \ac{gdn} and \ac{gsa} to achieve non trivial performance, while it did not help for \ac{bla} and \ac{swa}.
In addition for \ac{bla}, we introduce the token resolution as the quantity:
\begin{equation}
    \label{eq:tok_res}
    \frac{T}{2M-1}.
\end{equation}
We found that the performance on \ac{mqar} depends strongly on the token resolution (Fig.~\ref{fig:mqar}b), with a sharp maximum for a resolution of 2 tokens.
We hypothesize that it is due to the format of the task where key value pairs correspond to pairs of tokens.

Finally, we wanted to test whether \ac{bla} can model longer range dependencies than \ac{swa} for a given window size.
To test this, we increased the model dimension while matching the ``window sizes'' of \ac{bla} and \ac{swa} (respectively $2M-1$ and $w$).
We observed that the performance of \ac{bla} increased with higher model dimensions while \ac{swa}'s did not (Fig.~\ref{fig:mqar}c).
This suggests that even though keys and values end up overlapping due to the periodicity $T$, \ac{bla} can leverage orthogonality in the head dimension to enable retrieval beyond the window size.

\subsection{RegBench} \label{sec:regbench}

\begin{figure*}[ht]
  \vskip 0.2in
  \begin{center}
    \centerline{\includegraphics[width=0.6\textwidth]{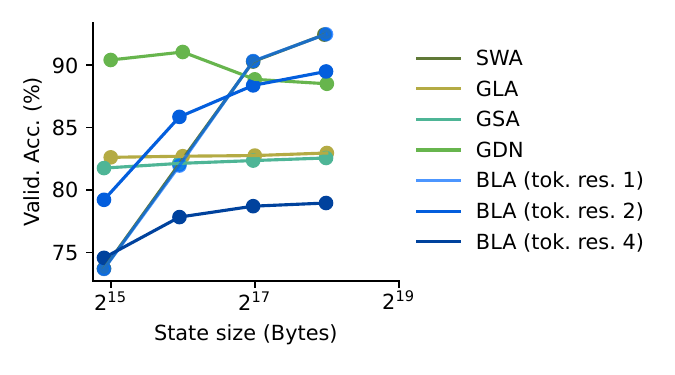}}
    \caption{
    Results on the RegBench task \citep{akyurek2024context} using 5000 DFAs.
    Accuracy of different models as the state size increases. We report the best test accuracy out of three different seeds. \ac{bla} in contrast to the other linear models increases its performance as the state size increases. Using a token resolution of two, \ac{bla} is both performing better for small state size and reaches similar performance to Full attention as the state size increases.
    }
    \label{fig:regbench}
  \end{center}
\end{figure*}

The RegBench benchmark is another task where linear architectures struggle to match the performance of Transformers \citep{von2025mesanet,akyurek2024context}. 
In this task, the objective is to infer the underlying structure of a grammar rule from a set of deterministic finite automata (DFAs). 
We train and evaluate different models on a dataset of 5,000 DFAs, including \acf{swa} with varying window sizes, \acf{gla}, \acf{gdn}, \acf{gsa}, and our proposed \ac{bla} model with state decay across different token resolutions (Eq.~\eqref{eq:tok_res}). 
To isolate the effect of the different \ac{ssm} modules, we run all models without the short 1D-convolution.

Our findings (Fig.~\ref{fig:regbench}) indicate that, in contrast to \ac{gla} and \ac{gdn}, \ac{bla} demonstrates a distinct state scaling advantage. 
As the state size increases—varying the number of modes for \ac{bla}, the number of slots for \ac{gsa}, the key expansion ratio for \ac{gla}, the value expansion ratio for \ac{gdn}, and the window size for \ac{swa}—our model's performance continues to improve. 
This suggests that \ac{bla}'s architecture is more effective at leveraging its state capacity on this benchmark. 
Of particular interest is the comparison with \ac{gsa}, where without the trigonometric interpolation kernel the model fails to increase it's performance as the number of slots increase.
Furthermore, consistent with observations on the \ac{mqar} task, we note a performance boost for \ac{bla} at smaller state sizes when the token resolution is set to 2, outperforming \ac{swa} in this regime.

\section{Discussion}
\label{sec:discussion}

We presented \acf{bla}, a novel linear attention model that can generate ``sharp'' attention masks and bridges the gap between \ac{swa} and \ac{abc}. 
We showed that our model can utilize its state size better than \ac{swa} particularly in the small state size regime and can scale the performance better than the other linear models with the increase of its state size.
The \ac{bla} model using an appropriate decay mechanism and period parameter can recover an exact implementation of \ac{swa}, and can therefore be understood as a more general version of \ac{swa}. 

A large body of work has been done to mitigate the quadratic complexity of full attention.
As explained in Section \ref{sec:bla_theory_ssm}, \ac{bla} is motivated by the early \ac{ssm} theory \cite{voelker2019legendre, gu2020hippo}.
However, while \acp{ssm} were introduced to compress continuous signals and involve a discretization step, \ac{bla} uses the theory from the standpoint of discrete interpolation, bypassing the need for discretization.
\ac{bla} is also related to the Attention with bounded memory control theory \cite{peng2022abc}, which keeps separated key and values states.
This choice however sacrifices some state efficiency, since to store $D$ KV associations of dimensions $D$, \acp{la} models need $D^2$ space while a model keeping KV separated needs $2D^2$ space.
We hypothesize that this difference explains why \ac{bla} does not fully match the efficiency of \acp{la} models on \ac{mqar} with roughly a factor of 2 (Fig.~\ref{fig:mqar}a).
We give a more complete discussion of related work in Appendix \ref{sec:app_related_work}.

As all methods, \ac{bla} has certain limitations and shortcomings that open up avenues for future research directions.
For example, \ac{bla} is sensitive to the choice of the period and number of modes, since depending the sequence length of the task at hand different hyper-parameters will give the optimal result. 
As a rule of thumb, we find the more number of modes the better, and the period should be chosen so we get a token resolution of 1 or 2.
Another limitation, is that the state capacity scales with $2D^2$ and not $D^2$ like other \ac{la}/\ac{ssm} models. 
One potential solution is instead of keeping two separate states for the keys and values, keep one latent representation for both, similar to multi-latent attention \cite{liu2024deepseek}.
Furthermore, given the impact of the token resolution on performance, designing a smarter interpolation mechanism seems like a promising direction for improving the model.

\newpage
\appendix
\onecolumn

\section{Related Work}
\label{sec:app_related_work}

A wide range of efficient attention models have emerged, each offering distinct strategies to scale linear architectures and compete with full attention. Similar to \ac{bla}, several approaches aim to enhance linear attention by enriching the feature map $\phi$, the decay/gating mechanism of the state update or by leveraging Fourier theory and complex-valued representations.
Beyond the \ac{la} and \ac{ssm} literature, \ac{bla}, similar to \ac{abc} and \ac{gsa}, can also be viewed as a method for compressing the key-value (KV) cache, aligning it with a number of efficient attention mechanisms.

Performer \cite{choromanski2020rethinking} and Random Feature Attention \cite{peng2021random} approximate the softmax kernel with random feature maps functions.
Hedgehog \cite{zhang2024hedgehog} uses a learnable MLP as a feature map to generate sharper attention mask and improve retrieval.
The Based architecture \cite{arora2024simple} uses a Taylor expansion feature map which improves retrieval but expands the head dimension significantly.
Gated Linear Attention, similar to Mamba and Mamba2 \cite{gu2023mamba,dao2024transformers}, uses simpler feature map, and adds data and feature dependent decay to Linear Attention \cite{yang2023gated}.
\citet{schlag2021linear} introduces the Delta rule to pack the state of Linear attention more efficiently.
Gated DeltaNet added gating to the Delta rule \cite{yang2024gated,yang2024parallelizing}, and is one of the leading linear models \cite{team2025kimi}.
Models built on \ac{la} can be formulated in terms of test-time training, where the key-value association can be viewed as an online learning objective \cite{von2023transformers}.
MesaNet \cite{von2023uncovering, von2025mesanet} makes this online learning objective depend on the whole trajectory to derive performance improvements. 

Similar to \ac{bla}, many models use the Fourier theory and more generally complex numbers to improve sequence modeling.
Fourier recurrent units \cite{zhang2018learning} summarizes the recurrent states along the temporal dimension with Fourier basis functions.
Rotational unit of memory \cite{dangovski2019rotational} uses unitary matrix to mitigate vanishing gradients.
FNet replaces the attention with a Fourier transform to mix the tokens \cite{lee2021fnet}.
The linear recurrent unit \cite{orvieto2023resurrecting} uses carefully initialized complex diagonal to model long range dependencies.
Megalodon \cite{ma2024megalodon} introduces complex exponential moving average to design powerful linear models.
CosFormer uses modulated cosine and sine states to add locality bias in Linear Attention \cite{qin2022cosformer}.

Finally, \ac{bla} is also related to works using Fourier theory to compress the KV cache.
A few examples include FAEDKV, which compresses the KV cache into the frequency domain using an Infinite-Window Fourier Transform \cite{li2025faedkv}, while DCT-Former uses the discrete cosine transform to compress the sequence and reduce the complexity of attention \cite{scribano2023dct}.
Those methods differ from \ac{bla} because they compress not only the keys and values but also the query sequence.

\section{Proofs}
\label{sec:app_proofs}

\textbf{State formula.} We prove here by recursion 
the closed from formulas for the states.
\begin{equation}
  \begin{split}
  \Kcos[m,t] & = \sum_{t'=1}^t \cosmw[(t-t')]~ \key_{t'},  \\
  \Ksin[m,t] & = \sum_{t'=1}^t \sinmw[(t-t')]~ \key_{t'}.
  \end{split}
\end{equation}
The state equations are true for $t=0$ as empty sums are zero.
We then substitute the formula and use trigonometric identities.
Then we have for $\Kcos[m,t+1]$:
\begin{align}
\Kcos[m, t+1] &= \cosmw \Kcos[m,t] - \sinmw \Ksin[m,t] + \key_{t+1} \nonumber \\
&= \cosmw \sum_{t'=1}^t \cosmw[(t-t')] \key_{t'}  - \sinmw \sum_{t'=1}^t \sinmw[(t-t')] ~\key_{t'} + \key_{t+1} \nonumber \\
&= \sum_{t'=1}^t \left(\cosmw \cosmw[(t-t')] - \sinmw \sinmw[(t-t')] \right) ~\key_{t'} + \key_{t+1} \nonumber \\
&= \sum_{t'=1}^t \cosmw[(t+1-t')] ~\key_{t'} + \key_{t+1} \nonumber \\
&= \sum_{t'=1}^{t+1} \cosmw[(t+1-t')] ~ \key_{t'}
\end{align}
because the last cosine term is one.
Similarly for $\Ksin[m,t+1]$: 
\begin{align}
\Ksin[m,t+1] &= \sinmw \Kcos[m,t] + \cosmw \Ksin[m,t] \nonumber \\
&= \sinmw \sum_{t'=1}^t \cosmw[(t-t')]~ \key_{t'} + \cosmw \sum_{t'=1}^t \sinmw[(t-t')] ~ \key_{t'} \nonumber \\
&= \sum_{t'=1}^t \left( \sinmw  \cosmw[(t-t')]  + \cosmw  \sinmw[(t-t')] \right) ~ \key_{t'} \nonumber \\
&= \sum_{t'=1}^{t} \sinmw[(t+1-t')] ~ \key_{t'} \\
&= \sum_{t'=1}^{t+1} \sinmw[(t+1-t')] ~ \key_{t'}
\end{align}
as the last sine term is 0.

\textbf{Fourier coefficients of translated Dirichlet kernels.} Consider the 
Dirichlet Kernel for an integer $M \geq 1$, 
\begin{align}
  D_{M}(t) = \frac{1}{2M-1} + \frac{2}{2M-1}\sum_{m=1}^{M-1} \cos \left(\frac{2\pi m t}{2M-1} \right)
\end{align}
Then $D_M$ is such that $D_M (0) =1$ , $D_M (t) = 0$ for $t \in [1, ... 2M-2]$  
We can also translate $D_M$ by an integer time $\Delta t \in [0, 2m-2]$ 
\begin{align}
D_{M}(t-\Delta t) &= \frac{1}{2M-1} + \frac{2}{2M-1}\sum_{m=1}^{M-1} \cos \left(\frac{2\pi m (t-\Delta t)}{2M-1} \right) \\
D_{M}(t-\Delta t) &= \frac{1}{2M-1} + \frac{2}{2M-1}\sum_{m=1}^{M-1} \cos \left(\frac{2\pi m t}{2M-1} - \frac{2\pi m \Delta t}{2M-1} \right) \\
&= \frac{1}{2M-1} + \frac{2}{2M-1}\sum_{m=1}^{M-1} \cos \left(\frac{2\pi m t}{2M-1}\right) \cos \left(\frac{2\pi m \Delta t}{2M-1}\right) \nonumber \\
&\quad +\frac{2}{2M-1}\sum_{m=1}^{M-1} \sin \left(\frac{2\pi m t}{2M-1}\right) \sin \left(\frac{2\pi m \Delta t}{2M-1}\right)
\end{align}

\section{Pseudo Code}
\label{sec:app_pseudo_code}

\begin{algorithm}[t]
\caption{Naive recurrent \ac{bla}}
\label{alg:recurrent_bla_ssm}
\begin{algorithmic}
\STATE {\bfseries Input:} $\Query, \Key, \Val \in \mathbb{R}^{\seqlen \times \modeldim}$, period $T \in \N$,
interpolation matrices ${\bf{A}} \in \mathbb{R}^{(2\nummodes-1) \times \nummodes}, {\bf{B}} \in \mathbb{R}^{(2\nummodes-1) \times \nummodes}$
\STATE Compute dilated time grid:
\STATE $\mathrm{dilated\_time} \gets \mathrm{round}\left(\frac{[0,\dots,2\nummodes-2]}{2\nummodes-1} \otimes T\right)$
\STATE $\omega \gets 2\pi/T$
\STATE Compute trigonometric components:
\STATE $\mathrm{modes} \gets \omega \otimes [0, 1, \dots, \nummodes-1]$
\STATE $\cos \gets \cos(\mathrm{modes})$
\STATE $\sin \gets \sin(\mathrm{modes})$ 
\STATE Initialize the KV states:
\STATE ${\bf{K}}^{\cos}, {\bf{K}}^{\sin}, {\bf{V}}^{\cos}, {\bf{V}}^{\sin} \gets {\bf{0}}^{\modeldim \times \nummodes}$ 
\STATE Initialize output:
\STATE ${\bf{O}} \gets {\bf{0}}^{\seqlen \times \modeldim}$

\FOR{$t = 0$ to $\seqlen-1$}
    \STATE Update the KV state (extra variables omitted):
    \STATE \quad $\bf{K}^{\cos} \gets (\cos \otimes \bf{K}^{\cos}) - (\sin \otimes \bf{K}^{\sin}) + \Key[t]$
    \STATE \quad $\bf{K}^{\sin} \gets (\sin \otimes \bf{K}^{\cos}) + (\cos \otimes \bf{K}^{\sin})$ 
    \STATE \quad $\bf{V}^{\cos} \gets (\cos \otimes \bf{V}^{\cos}) - (\sin \otimes \bf{V}^{\sin}) + \Val[t]$
    \STATE \quad $\bf{V}^{\sin} \gets (\sin \otimes \bf{V}^{\cos}) + (\cos \otimes \bf{V}^{\sin})$

    \STATE Compute interpolated keys and values:
    \STATE \quad $\tilde{\bf{K}} \gets {\bf{A}} \cdot {\bf{K}}^{\cos} + {\bf{B}} \cdot {\bf{K}}^{\sin} \quad \in \R^{\modeldim \times (2\nummodes-1)}$
    \STATE \quad $\tilde{\bf{V}} \gets {\bf{A}} \cdot {\bf{V}}^{\cos} + {\bf{B}} \cdot {\bf{V}}^{\sin}$

    \STATE Update output:
    \STATE \quad ${\bf{M}}_{\mathrm{BLA}} \gets \mathrm{where}(t \geq \mathrm{dilated\_time}, 0.0, -\infty)$
    \STATE \quad ${\bf{O}}[t] \gets \mathrm{Softmax}\left(\frac{\Query[t] \cdot \tilde{\bf{K}}^\top}{\sqrt{D}} + \bf{M}_{\mathrm{BLA}}\right) \cdot \tilde{\bf{V}}$
\ENDFOR
\STATE {\bfseries Return:} $\bf{O}$
\end{algorithmic}
\end{algorithm}

\begin{algorithm}[tb]
\caption{Efficient chunk \ac{bla}}
\label{alg:chunk_bla_swa}
\begin{algorithmic}
\STATE {\bfseries Input:} $\Query, \Key, \Val \in \mathbb{R}^{\seqlen \times \numheads \times \modeldim}$, period $T \in \mathbb{R}^{\numheads}$,
interpolation matrices ${\bf{A}} \in \mathbb{R}^{2\nummodes-1 \times \nummodes}, {\bf{B}} \in \mathbb{R}^{2\nummodes-1 \times \nummodes}$, chunk size $C \in [L]$
\STATE Clamp period: $T \gets \max(T, 2 \cdot \nummodes - 1)$
\STATE Compute dilated time grid:
\STATE $\mathrm{dilated\_time} \gets \mathrm{round}\left(\frac{[0,\dots,2\nummodes-1-1]}{2\nummodes-1} \otimes T\right)$
\STATE $\omega \gets 2\pi/T$
\STATE Compute modes: $\mathrm{modes} \gets \omega \otimes [0, 1, \dots, \nummodes-1]$
\STATE Initialize output: ${\bf{O}} \gets {\bf{0}}^{\seqlen \times \numheads \times \modeldim}$
\STATE Divide $\bf{\Query}, \bf{\Key}, \bf{\Val}, \bf{O}$ into $N=\frac{\seqlen}{C}$ blocks $\{ \bf{\Query}_{[1]}\dots \bf{\Query}_{[N]}\}$, $\{ \bf{\Key}_{[1]}\dots \bf{\Key}_{[N]}\}$, $\{ \bf{\Val}_{[1]}\dots \bf{\Val}_{[N]}\}$, $\{ \bf{O}_{[1]}\dots \bf{O}_{[N]}\}$ of size $C\times\numheads\times\headdim$ each
\STATE Initialize compressed KV states:
\STATE ${\bf{\Key}}_{\mathrm{prev}}, {\bf{\Val}}_{\mathrm{prev}} \gets {\bf{0}}^{\numheads \times \modeldim \times 2\nummodes-1}$ 
\FOR{$n = 0$ to $N$}
\STATE Compute chunk interpolation coefficients:
\STATE \quad $t \gets [n,n+1,\dots,n+C-1]$
\STATE \quad $\mathrm{interpolate} \gets \cos(t \cdot \mathrm{modes}) \cdot {\bf{A}}^\top + \sin(t \cdot \mathrm{modes}) \cdot {\bf{B}}^\top \in \mathbb{R}^{C\times \numheads \times 2\nummodes-1}$ 
\STATE Update compressed KV states:
\STATE \quad ${\tilde{\bf{\Key}}} \gets \Key_{[n]} \otimes \mathrm{interpolate}$
\STATE \quad ${\tilde{\bf{\Key}}} \gets \mathrm{cumsum}(\tilde{\bf{\Key}})$ over $C$
\STATE \quad ${\tilde{\bf{\Val}}} \gets \Val_{[n]} \otimes \mathrm{interpolate}$
\STATE \quad ${\tilde{\bf{\Val}}} \gets \mathrm{cumsum}(\tilde{\bf{\Val}})$ over $C$
\STATE \quad ${\tilde{\bf{\Key}}} \gets {\bf{\Key}}_{\mathrm{prev}} + \tilde{\bf{\Key}}$
\STATE \quad ${\tilde{\bf{\Val}}} \gets {\bf{\Val}}_{\mathrm{prev}} + \tilde{\bf{\Val}}$
\STATE \quad ${\bf{K}}_{\mathrm{prev}} \gets {\tilde{\bf{K}}}[C-1]$ 
\STATE \quad ${\bf{V}}_{\mathrm{prev}} \gets {\tilde{\bf{V}}}[C-1]$
\STATE Compute attention weights:
\STATE \quad ${\bf{M}}_{\mathrm{BLA}} \gets \mathrm{where}(t  \geq \mathrm{dilated\_time}, 0.0, -\infty)$
\STATE \quad ${\bf{O}}_{[n]} \gets  \mathrm{Softmax}\left(\frac{\Query_{[n]} \cdot {\tilde{\bf{K}}}}{\sqrt{\modeldim}} + {\bf{M}}_{\mathrm{BLA}}\right) \cdot {\tilde{\bf{V}}}$
\ENDFOR
\STATE {\bfseries Return:} $\bf{O}$
\end{algorithmic}
\end{algorithm}

\section{Hyperparameters}
\label{sec:app_hparams}

\subsection{MQAR experiment}

\begin{table}[t]
  \caption{The hyperparameters used for the MQAR experiment}
  \label{tab:mqar_hparams}
  \begin{center}
    \begin{small}
      \begin{tabular}{lcccr}
        \toprule
        Hyperparameter      & Values     \\ 
        \midrule
        Batch size          & $32$    \\ 
        Optimizer           & AdamW    \\
        Learning rates      & 1e-4, 3e-4, 1e-3, 3e-3    \\
        Weight decay        & 1e-2 \\
        \quad LR decay      & Cosine    \\
        \quad min. LR       & 0    \\
        \quad Warm-up steps & 10\%    \\
        Num. epochs         & $32$   \\
        Model dimensions    & 32, 64, 128, 256   \\
        Num. heads          & 1, 2   \\
        Num. layers         & 2   \\
        Expand key dim.     & 0.5, 1, 2   \\
        Vocab. size         & $8,192$    \\
        Train dataset size  & $100,000$    \\
        Val. dataset size   & $3,000$    \\
        Sequence length     & $512$  \\
        Num. KV pairs       & $64$  \\
        \midrule
        \ac{swa} window sizes & $\log_2(w) \in [3,...,8]$    \\
        \ac{gsa} num. slots   & $\log_2(S+1) \in [3, ..., 7]$  \\
        \ac{bla} num. modes   & $\log_2(M) \in [2, ..., 6]$  \\
        \ac{bla} period       & $T = 2(2M-1)$  \\
        \bottomrule
      \end{tabular}
    \end{small}
  \end{center}
  \vskip -0.1in
\end{table}

\subsection{RegBench experiment}

Similar to \cite{von2025mesanet,akyurek2024context}, we train models across a small search-space and in the paper we show the network with the best validation performance across three different network initializations. As we mention in the main text, we remove the short convolution from all of our models, and we use only the RegBench dataset with $5000$ DFAs. For more details see Table \ref{tab:regbench_hparams}.
 
\begin{table}[t]
  \caption{The hyperparameters used for the RegBench experiment}
  \label{tab:regbench_hparams}
  \begin{center}
    \begin{small}
      \begin{tabular}{lcccr}
        \toprule
        Hyperparameter    & Search     \\ 
        \midrule
        Batch size        & $32$    \\ 
        Number of epochs  & $60$    \\
        Model dimensions  & $128$    \\
        Number of layers  & $4$    \\
        Number of heads   & $4$    \\
        Optimizer         & AdamW    \\
        \quad Learning rate  & $0.0001$, $0.0003$, $0.001$    \\
        \quad Weight decay   & $0.$,$0.01$, $0.5$  \\
        Scheduler            & Cosine with Warmup \\
        \quad Minimum l. r.  & 0 \\ 
        \quad Warm-up steps  & 2000 \\
        \midrule
        \ac{swa} window sizes  & $\log_2(w) \in [3,...,6]$    \\
        \ac{bla} num. modes    & $\log_2(M) \in [3, ..., 6]$  \\
        \ac{gsa} num. slots    & $[15, ..., 127]$  \\
        \quad \ac{bla} period        & $T = [1, 2, 4](2M-1)$  \\
        \ac{gla} expand keys   & $[1, 2, 4, 8]$  \\
        \ac{gdn} expand values & $[1, 2, 4, 8]$  \\
        \bottomrule
      \end{tabular}
    \end{small}
  \end{center}
  \vskip -0.1in
\end{table}

\end{document}